\documentclass{article}

\usepackage[left=3.2cm, right=3.2cm, top=2.5cm, bottom=2.5cm]{geometry}
\usepackage{mathpazo}

\usepackage[utf8]{inputenc}   
\usepackage[T1]{fontenc}      
\usepackage{amsfonts}         
\usepackage{amsmath, amssymb, amsthm, mathtools}  

\usepackage[numbers]{natbib}
\usepackage[colorlinks=true, citecolor=BrickRed, linkcolor=RoyalBlue, bookmarks=false, pagebackref=true, breaklinks=true]{hyperref}
\usepackage{url}              

\usepackage{booktabs}         
\usepackage[table,usenames,dvipsnames]{xcolor}  
\usepackage{graphicx}         
\usepackage{subcaption}       
\usepackage{wrapfig}          
\usepackage{arydshln}         

\usepackage{nicefrac}         
\usepackage{microtype}        
\usepackage{caption}
\usepackage{bm}               
\usepackage{dsfont}           
\usepackage{multirow}         
\usepackage{multicol}         
\usepackage{array}            
\usepackage{bbding}           
\usepackage{lipsum}           
\usepackage{verbatim}         
\usepackage{pifont}           
\usepackage{enumitem}         
\usepackage{algorithm}
\usepackage{algorithmic}
\usepackage{tcolorbox}        
\usepackage[capitalize,noabbrev]{cleveref} 

\usepackage{framed}           
\usepackage{colortbl}         
\usepackage[textsize=tiny]{todonotes}  

\definecolor{Gray}{gray}{0.9}
\definecolor{BrickRed}{rgb}{0.6,0,0}
\definecolor{RoyalBlue}{rgb}{0,0,0.8}
\definecolor{Tdgreen}{rgb}{0,0.4,0.7}
\hypersetup{citecolor=BrickRed, linkcolor=RoyalBlue}

\definecolor{mycolor}{rgb}{0.92,0.92,0.98}
\definecolor{color1}{rgb}{0.86, 0.72, 1}
\definecolor{color2}{rgb}{0.89, 0.78, 1}
\definecolor{color3}{rgb}{0.92, 0.85, 1}
\definecolor{color4}{rgb}{0.66, 0.815, 0.62}
\definecolor{color5}{rgb}{0.77, 0.87, 0.70}
\definecolor{color6}{rgb}{0.88, 0.93, 0.85}
\definecolor{mygray}{gray}{.93}

\title{TrustLoRA: Low-Rank Adaptation for Failure Detection under Out-of-distribution Data}
\author{Fei Zhu\textsuperscript{\rm 1}, Zhaoxiang Zhang\textsuperscript{\rm 2,3}\\
\\
\normalsize \textsuperscript{\rm 1} CAIR-HKISI, Chinese Academy of Sciences, HongKong, China\\
\normalsize \textsuperscript{\rm 2} MAIS, Institute of Automation, Chinese Academy of Sciences, Beijing 100190, China\\
\normalsize \textsuperscript{\rm 3} School of Artificial Intelligence, UCAS, Beijing, 100049, China\\
}
\date{}

\begin{document}

\maketitle

\begin{abstract}
		Reliable prediction is an essential requirement for deep neural models that are deployed in open environments, where both covariate and semantic out-of-distribution (OOD) data arise naturally. In practice, to make safe decisions, a reliable model should accept correctly recognized inputs while rejecting both those misclassified covariate-shifted and semantic-shifted examples. Besides, considering the potential existing trade-off between rejecting different failure cases, more convenient, controllable, and flexible failure detection approaches are needed. To meet the above requirements, we propose a simple failure detection framework to unify and facilitate classification with rejection under both covariate and semantic shifts. Our key insight is that by separating and consolidating failure-specific reliability knowledge with low-rank adapters and then integrating them, we can enhance the failure detection ability effectively and flexibly. Extensive experiments demonstrate the superiority of our framework.
		\end{abstract}

	\section{Introduction}\label{sec:introduction}
\begin{figure*}[t]
		\begin{center}	
	\centerline{\includegraphics[width=\textwidth]{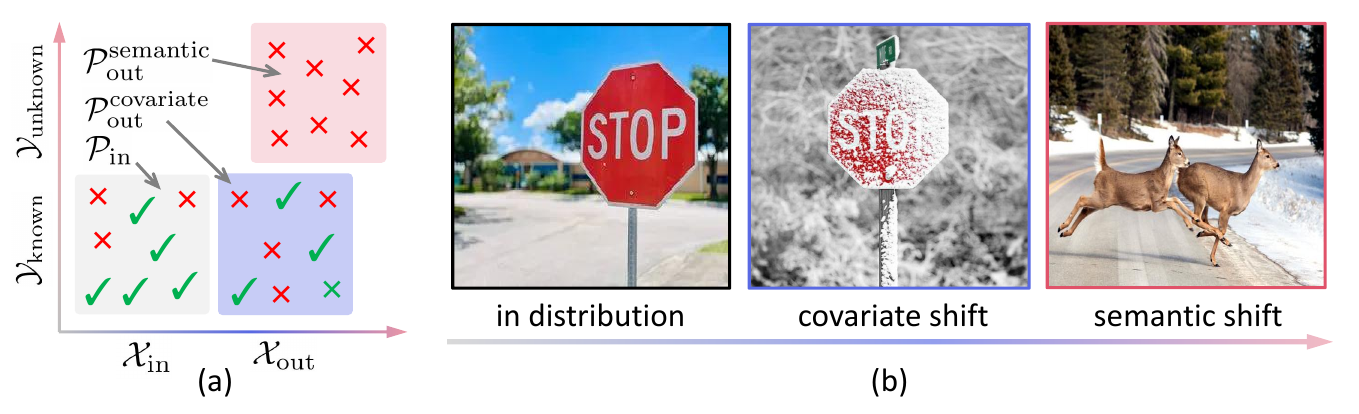}}
			\vskip -0.1 in
			\caption{(a) Failure detection rejects both the (\textcolor{red}{\ding{55}}) misclassified covariate-shifted and all semantic-shifted OOD samples, and accepts the (\textcolor{green}{\ding{51}}) correct prediction. (b) Illustration of three types of common failure cases in the natural open environment.}
			\label{figure-1}
		\end{center}
		\vskip -0.2 in
	\end{figure*}

	Deep neural models have achieved remarkable performance in closed-world scenarios, assuming that train and test sets come from the same distribution. However, in practice, out-of-distribution (OOD) data naturally arises during the deployment \cite{zhang2025unsupervised, zhang2025learning, zhu2024open}, which mainly includes two types named \emph{covariate shifts} and \emph{semantic shifts} \cite{bai2023feed}. Specifically, as
	depicted in Fig.~\ref{figure-1}, a model trained on in-distribution (ID) data may encounter covariate shifts such as conditions with snowy night or corrupted inputs resulting from camera noise and sensor degradation \cite{hendrycks2018benchmarking}. Unfortunately, the model often suffers significant performance
	deterioration when deployed in those scenarios. To ensure safety, it is expected to reject wrong predictions instead of accepting them blindly. Alternatively, unknown categories with semantic shifts may also emerge \cite{hendrycks2018deep}. In this case, the model must learn to reject to make incorrect decisions.
	
	In recent years, both covariate and semantic shifts have received extensive attention, and have been formulated as OOD generalization \cite{wang2024discovering, liu2025iw} and detection \cite{hendrycks2018deep, ma2025towards, wei2025fine, jia2025enhancing, zhang2025unsupervised, cheng2023average} problems, respectively. Concretely, the former focuses on recognizing inputs with covariate shifts while the latter focuses on rejecting inputs with semantic shifts. Instead of pursuing those two problems independently, \cite{bai2023feed} handles OOD generalization and detection simultaneously by leveraging unlabeled wild data consisting of both covariate and semantic shifts during training. However, the aforementioned efforts still have primary limitations.
	First, for OOD generalization, there is no rejection option involved, and accepting misclassified covariate-shifted inputs could lead to catastrophic issues. Second, for OOD detection, the performance of prevalent methods drops a lot when facing covariate shifts, and rejecting semantic-shifted samples while accepting all covariate-shifted samples may also lead to serious safety issues.

	In addition, the trade-off between the rejection of different failure sources further complicates the problem. There are a few studies \cite{zhu2024rcl, zhu2022learning, zhu2023openmix, li2024sure} focused on developing reliable models that can reject both misclassified ID and semantic-shifted OOD data. Nevertheless, they typically overlook covariate-shifted samples, and it is hard to distinguish correct covariate-shifted samples from semantic-shifted ones. Besides, \textbf{\emph{different failure sources are not always predefined and can emerge continually}} in practical scenarios. For instance, an autonomous
	driving system performs classification with misclassification rejection on ID data under a normal environment (e.g., clean inputs on a sunny day), and switches to more challenging failure detection under covariate shifts when facing sensory degeneration or bad weather (Fig.~\ref{figure-1} (b, Middle)). Moreover, when a car drives into the countryside, it may encounter unexpected novel objects such as sheep and deer (Fig.~\ref{figure-1} (b, Right)), where the model should perform OOD detection and make a warning. Therefore, a single prefixed, static solution lacks the flexibility to explore and calibrate the trade-off among different requirements, and there is a demand for developing \textbf{\emph{flexible}} and \textbf{\emph{controllable}} failure detection methods.
	
	The goal of this paper is to show that the above-mentioned limitations and requirements can be considerably addressed. For one thing, we aim to predict and accept correctly classified covariate-shifted examples while rejecting those misclassified ones and all unknown samples with semantic shifts. As illustrated in Fig.~\ref{figure-1} (a), unlike the OOD detection problem that defines ``positive'' and ``negative'' with regard to the label space, failure detection directly specifies the distinction by the correctness of model’s predictions, which is more reasonable and aligned with the requirement in practical applications. 
	For another, considering the trade-off between rejecting different failure sources, we aim to develop a more flexible method that enables us to easily separate, consolidate, and incorporate different reliable knowledge regarding surrounding environments. 
	
	\textbf{Contributions.} (1) We study the failure detection problem under both covariate and semantic shifts, and call for flexible and controllable methods for reliability enhancement. (2) We propose a reliability arithmetic framework with low-rank adapters to compress and consolidate reliability knowledge effectively and flexibly. To the best of our knowledge, this work is the first to separate and compress reliability knowledge via low-rank adapters. Further, a random projection strategy is proposed for rank adaptation to enhance the tuning efficiency. (3) Comprehensive experiments demonstrating the strong performance of our method, as well as the flexibility of reliability edition. 
	
\section{Related Work}
Current out-of-distribution detection methods have been proposed under the setting of post hoc or training regularization, aiming to reject samples from unseen classes. Some post hoc methods \cite{liu2020energy, sun2022out, park2023nearest, kim2023neural, liu2023gen} focus on designing proper confidence scores, which others \cite{djurisic2022extremely, djurisic2022extremely} remove undesirable parts of feature or activation to facilitate the separation of ID and OOD examples. Training regularization approaches \cite{hendrycks2018deep, ma2025towards, wei2025fine, cheng2024breaking, jia2025enhancing} often require real or synthesized auxiliary dataset with extra training processes. For instance, OE \cite{hendrycks2018deep} leverages auxiliary outliers the enhance semantic OOD detection. Cheng et al., \cite{cheng2024breaking} studied whether and how expanding the training set using generated data can improve reliable prediction.
Some recent work \cite{zeng2025local, zeng2025towards} also investigates the OOD detection of vision-language models.
Nevertheless, current methods could harm the performance of detecting misclassified examples from known classes. 
Bai et al., \cite{bai2023feed} leveraged unlabeled wild data consisting of covariate and semantic shifts to build a model to recognize covariate-shifted data while rejecting semantic-shifted data. However, there is no rejection option in covariate shifts generalization, and users would accept the widely existing misclassification blindly. 
Contrary to the above prior works, in this work, we aim to reject misclassified covariate-shifted samples reliably.

Recently, there are a few studies \cite{zhu2024rcl, cen2023devil, zhu2023openmix, li2024sure} focused on developing reliable models that can reject both misclassified ID and semantic-shifted OOD data. For example, Zhu et al., \cite{zhu2023openmix, zhu2024rcl} observed that existing popular OOD detection methods are harmful for misclassification detection on clean ID test data, and proposed unified failure detection methods by exploring outlier data \cite{zhu2023openmix} or reliable continual learning paradigm \cite{zhu2024rcl}. Cen et al., \cite{cen2023devil} found that the uncertainty distribution of wrongly classified samples is extremely close to semantic-shifted samples rather than known and correctly classified samples, and proposed FS-KNN, which is an improvement of the KNN score. Li et al., \cite{li2024sure} proposed a method named SURE for reliable prediction by combining multiple techniques, across model regularization, classifier and optimization.
Nevertheless, they typically overlook covariate-shifted samples, and it is hard to distinguish correct covariate-shifted samples from semantic-shifted ones. Besides, they often train a model from scratch or fully fine-tune it, which is computationally heavy and inefficient.

	\section{Problem Formulation}
	
	\textbf{Training on in-distribution data.}
	We focus on the multi-class classification setting. Let $\mathcal{X} \subset \mathbb{R}^d$ be an input space, $\mathcal{Y} = [K] =  \{1, . . . , K\}$ denotes the label space and $\mathcal{P}_{\rm in}$ be the underlying in-distribution (ID) over $\mathcal{X} \times \mathcal{Y}$.  Given a labeled training set $\mathcal{D}^{\rm train}_{\rm in}=\{(\mathbf{x}_i, y_i)\}^N_{i=1}$ comprising $N$ samples drawn \textit{i.i.d.} from the joint data distribution $\mathcal{P}_{\rm in}$, multi-class classification aims to learn a classifier $h: \mathcal{X} \rightarrow \mathcal{Y}$ with low misclassification error. Typically, we learn a function $f: \mathcal{X} \rightarrow \mathbb{R}^K$ that 
	yields the posterior distributions of a given input
	by minimizing an empirical surrogate risk, e.g., cross-entropy (CE) loss, on $\mathcal{D}^{\rm train}_{\rm in}$, and then $h(\mathbf{x}) = \arg\max_{y \in [K]}f_y(\mathbf{x})$.
	
	\textbf{Inference in open environments with wild data.} Trained on the ID data, a classifier
	$f$ deployed in open environments can encounter various out-of-distribution (OOD) shifts, as shown in Figure 1(a). The OOD data can be grouped into covariate and semantic shifts:
	1) Covariate shifts $\mathcal{P}^{\rm cov}_{\rm out}$ has the same label space $\mathcal{Y}$ as the training data, but the input space $\mathcal{X}^{\rm cov} \subset \mathbb{R}^d$ undergoes shifting.
	2) Semantic shifts $\mathcal{P}^{\rm sem}_{\rm out}$ represents new-class shifted samples that do not belong to any known classes, i.e., $y \notin \mathcal{Y}$. We further assume that the input space $\mathcal{X}^{\rm sem}$ and $\mathcal{X}$ are also in different subsets of $\mathbb{R}^d$, which makes OOD detection possible.
	For inference with covariate shifts, existing literature formulates the OOD generalization problem to improve the classification accuracy of covariate-shifted samples. For semantic shifts, prior studies formulate the OOD detection problem \cite{hendrycks2018deep} to separate ID and semantic OOD.
	
	\textbf{Formulation of failure detection in the wild.}  In practice, one is likely to encounter both types of
	samples during classifier deployment. To this end, failure detection allows for abstention on both misclassified covariate-shifted and semantic-shifted data, while only accepting correctly classified inputs from known classes ($y = h(\mathbf{x}) ~{\rm and}~ y \in \mathcal{Y}$). 
	Formally, the goal of failure detection is to learn the classifier $h$ and design a rejector $r: \mathbb{R}^d \rightarrow \{0, 1\}$, where an ideal rejector can ensure to make safe decisions by separating correctly classified samples from misclassified ones or semantic OOD data:
	\begin{equation}\label{fd}
		r(\mathbf{x}) = \left\{ 
		\begin{aligned}
			&1,~~\mathbf{x} \in [\mathcal{P}_{\rm in} \cup \mathcal{P}^{\rm cov}_{\rm out}](y \neq h(\mathbf{x})) \cup \mathcal{P}^{\rm sem}_{\rm out} \\
			&0,~~\mathbf{x} \in [\mathcal{P}_{\rm in} \cup \mathcal{P}^{\rm cov}_{\rm out}] (y = h(\mathbf{x}))
		\end{aligned}
		\right..
	\end{equation}
	The distinction between those three problems is as follows: OOD generalization only focuses on classification accuracy and has \textbf{\emph{no rejection option}}; OOD detection \textbf{\emph{only rejects}} semantic-shifted samples from unknown classes ($y \notin \mathcal{Y}$), and blindly accepts misclassified samples from known classes ($y \neq h(\mathbf{x}) ~{\rm and}~ y \in \mathcal{Y}$). Besides, misclassification detection (MisD) focuses on known classes and rejects misclassified ones. Failure detection provides a unified classification with rejection framework that satisfies the practical requirements.  

	\begin{figure}[h]
		\begin{center}
			\centerline{\includegraphics[width=0.9\textwidth]{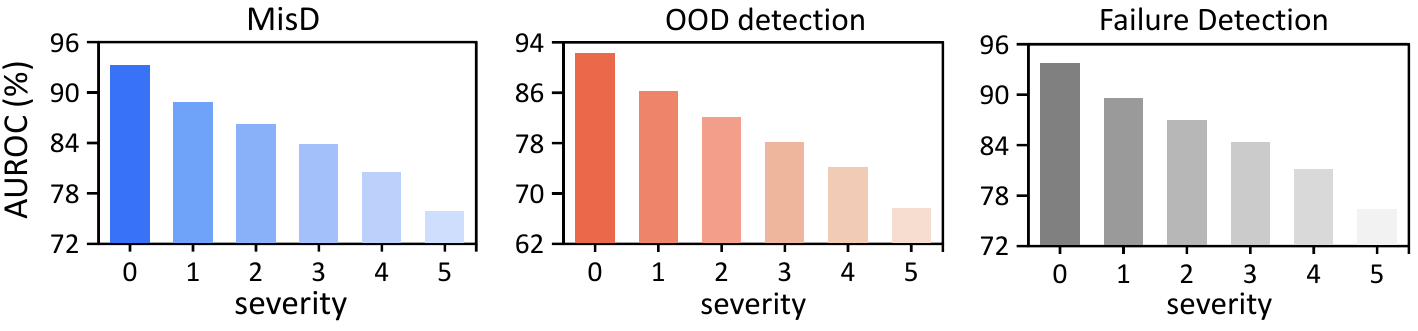}}
			\vskip -0.15 in
			\caption{Covariate shifts complicate failure detection.}
			\label{figure-2}
		\end{center}
		\vskip -0.1 in
	\end{figure}
	\textbf{Failure detection in the wild is challenging.} Prior works study the rejection without considering covariate shifts that will be anticipated at inference time. 
	Actually, failure detection under covariate shifts is quite difficult. As shown in Fig.~\ref{figure-2} (ResNet-18  \cite{he2016deep} trained on CIFAR-10): (1) Within known classes, covariate shifts make it much harder to separate misclassified examples from correct ones. When increasing the corruption severity, the performance of MisD continually drops. (2) Considering OOD detection performance, the model struggles to distinguish between known and unseen classes when the samples of known classes undergo covariate shifts. (3) From (1)-(2), we know that $\mathcal{P}^{\rm cov}_{\rm out}(y \neq h(\mathbf{x}))$ and $\mathcal{P}^{\rm cov}_{\rm out}(y = h(\mathbf{x}))$ are hard to be separated, and the confidence distributions of $\mathcal{P}^{\rm cov}_{\rm out}$ and $\mathcal{P}^{\rm sem}_{\rm out}$ are also mixed.
	
	\section{The Proposed Framework: TrustLoRA}

	\begin{figure*}[t]
		\begin{center}
			\vskip -0.05 in
			\centerline{\includegraphics[width=\textwidth]{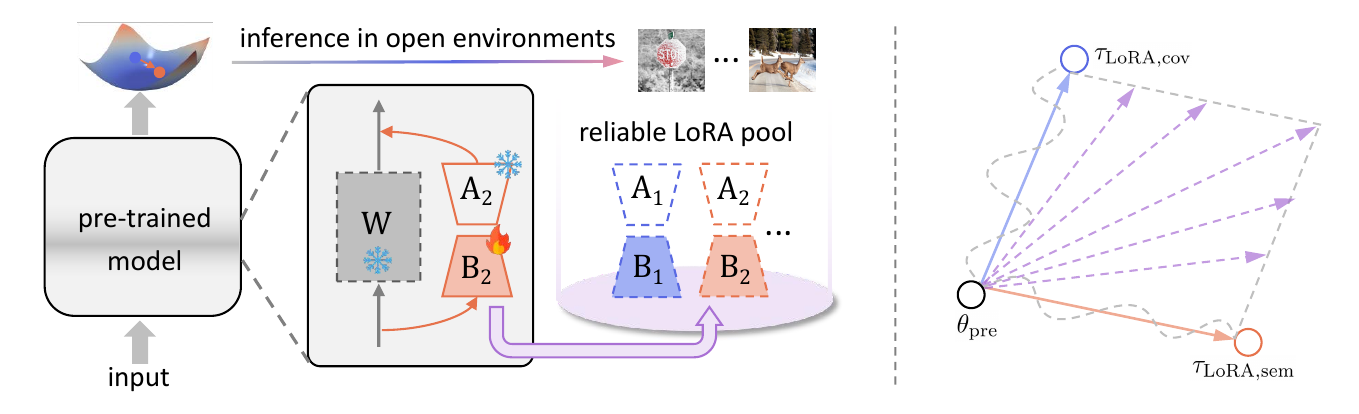}}
			\vskip -0.1 in
			\caption{Illustration of the proposed reliability arithmetic framework. (Left) We freeze the pre-trained backbone and add a LoRA module to acquire failure-specific knowledge. (Right) The LoRAs will be merged via arithmetic for unified failure detection in the wild.}
			\label{figure-3}
		\end{center}
		\vskip -0.15 in
	\end{figure*}
	\textbf{Limitation of failure-specific full training.} 
	The failure sources are rich in uncontrolled environments, while current methodologies predominantly focus solely on rejecting one specific failure case. This paradigm, however, has evident limitations: 
	\textbf{(1)} \textbf{\emph{Unrecoverable.}} Enhancing the ability of rejection on one specific failure may lead to unrecoverable damage on other aspects of the model, since it has been empirically revealed that trade-off existed when rejecting different failure sources \cite{jaeger2022call, kim2023unified}. This is undesirable in practice: an autonomous car can not return to its ``standard'' mode for normal environment after full tuning in OOD environment. 
	\textbf{(2)} \textbf{\emph{Inflexible.}} Full training with failure-specific optimization objectives often leads to a static solution. Considering the complexity of open environments, it is beneficial to have convenient ways that can flexibly adjust the trade-off at inference time without full retraining.
	\textbf{(3)} \textbf{\emph{Inefficient.}} When facing new failure cases, full training is computationally intensive and time-consuming. In practice, we expect the model to handle novel failure sources with minimal overhead in latency.
	
	\textbf{Reliability knowledge separation and integration.} With the above limitations in mind, we propose to develop failure detection framework with separable and combinable reliability knowledge, which is different remarkably from the prior efforts. As demonstrated by \cite{gueta2023knowledge}, knowledge can be represented by a region in weight space. Our high-level idea is to compress reliability knowledge regarding different failure cases and then selectively integrate them based on real-world requirements. To this end, two important questions arise: how to get failure-specific knowledge and how to compress it. 
	\textbf{(1)} \textbf{\emph{Acquire reliability}.} Many methods have been developed in recent years for reliable prediction, and they often excel at one specific failure case. 
	Those methods form a rich and diverse toolbox, which can be interpreted as encapsulating the specific reliability knowledge naturally.
	\textbf{(2)} \textbf{\emph{Compress reliability}.} Common strategies to compress knowledge such as pruning and knowledge distillation often suffer from the heavy computation issue, which conflicts with the efficient principle. Therefore, we hope to compress knowledge to a small set of parameters, enabling cheap computation and lightweight integration.

	
	\subsection{Reliability Separation with Low-Rank Adaptation}
	We propose a novel \textbf{TrustLoRA} framework to acquire and integrate trustworthy knowledge, which is illustrated in Fig.~\ref{figure-3} and detailed below. 
    
	\textbf{LoRA-adapted reliability} acquiring.  To acquire and compress specific reliable knowledge related to covariate shifts,  we propose to fine-tune the model in specific low-rank subspace. Concretely, we leverage parameter efficient tuning technique with an auxiliary low-rank adapter (LoRA) \cite{hu2021lora}. As illustrated in Fig.~\ref{figure-3}, LoRA composes of two rank decomposition matrices $\mathbf{B} \in \mathbb{R}^{u \times r}$ and $\mathbf{A} \in \mathbb{R}^{r \times v}$ where $r \in \mathbb{N}$ is the rank and $r \ll {\rm min}(u, v)$. $v$ and $u$ are the dimensionality of the input $\mathbf{\hat{x}} \in \mathbb{R}^{v}$ for current layer and hidden features, respectively. Therefore, $\mathbf{B}\mathbf{A} \in \mathbb{R}^{u \times v}$ has the same size as the parameters, i.e., $\mathbf{W} \in \mathbb{R}^{u \times v}$, of the corresponding fully-connected layer in the feature extractor. The modified forward pass with LoRA becomes:
	\begin{equation}
		\mathbf{z} = (\mathbf{W} + \mathbf{BA})\mathbf{\hat{x}} = \mathbf{W}\mathbf{\hat{x}} + \mathbf{BA}\mathbf{\hat{x}},
	\end{equation}
	where $\mathbf{z} \in \mathbb{R}^{u}$ is the output, which will be the input of the next layer after passing non-linear activation. During the training stage, the original parameters $\mathbf{W}$ remain frozen, while only $\mathbf{A}$ and $\mathbf{B}$ are trainable, which is low-cost and parameter efficient.

	To acquire and separate reliable knowledge in dynamic open environments, we propose to optimize the failure-specific objectives via the LoRA branch as follows. In this work, we follow most of existing studies that assume the real OOD data is unavailable. For covariate shifts, we leverage AugMix \cite{hendrycks2019augmix}, which is a simple augmentation method with the following learning objective:
	\begin{equation}\label{augmix}
		\begin{aligned}
			\mathcal{L}_{{\rm LoRA, cov}} = \mathcal{L}_{{\rm CE}}(f(\mathbf{x}), y) + \lambda {\rm JS}\left(f(\mathbf{x});f(\mathbf{x_{\rm aug1}});
			f(\mathbf{x_{\rm aug2}})\right).
		\end{aligned}
	\end{equation}
	Denote $\overline{f} = \left(f(\mathbf{x}) + f(\mathbf{x_{\rm aug1}}) + f(\mathbf{x_{\rm aug2}})\right)/3$ the averaged posterior distributions of $\mathbf{x}$ and its augmented variants, and then the JS loss is:
	$${\rm JS}\left(f(\mathbf{x});f(\mathbf{x_{\rm aug1}});f(\mathbf{x_{\rm aug2}})\right) = \frac{1}{3}\left(\mathcal{L}_{{\rm KL}}(f(\mathbf{x}), \overline{f}) + \mathcal{L}_{{\rm KL}}(f(\mathbf{x_{\rm aug1}}), \overline{f}) + \mathcal{L}_{{\rm KL}}(f(\mathbf{x_{\rm aug2}}), \overline{f})\right).$$ 
	For semantic shifts, we use OE \cite{hendrycks2018deep} to acquire the knowledge of unknown classes by introducing auxiliary outliers $\mathcal{D}_{\text{aux}}$. Specifically, we minimize the following objective:
	\begin{equation}\label{oe}
		\begin{aligned}
			\mathcal{L}_{{\rm LoRA, sem}} = 
			\mathcal{L}_{{\rm CE}}(f(\mathbf{x}), y) +  \lambda \cdot \mathcal{L}_{{\rm KL}}\left(f(\mathbf{x_{\rm aux}}), \mathcal{U}([K])\right),
		\end{aligned}
	\end{equation}
	where $\lambda > 0$ is a scalar, $\mathcal{U}([K])$ represents the uniform distribution over the training label space $\mathcal{Y} = [K]$. $\{\mathbf{A},\mathbf{B}\}$ denotes all trainable parameters. Since the pre-trained backbone is frozen, the newly added LoRA captures the residual knowledge regarding the specific learning objectives. We would like to clarify that we do not propose novel failure-specific learning objectives in this paper. Instead, we focus on designing a unified framework to integrate different sources of reliability knowledge in a flexible and parameter-efficient manner.

	\textbf{LoRA with random projection.}  
	The common way is to initialize $\mathbf{B}$ with an all-zero matrix, and initialize $\mathbf{A}$ with a normal distribution, where each element is independently sampled from a standard Gaussian distribution. In other words, LoRA first projects the input $\mathbf{\hat{x}}$ into a low-rank space via random projection, and then decodes it to the original space. For random projection, the Johnson-Lindenstrauss \cite{dasgupta2003elementary} states that the pairwise relation between any two data points can be preserved in an appropriate lower-rank space. Therefore, we fix the parameters of $\mathbf{A}$ once initialized and only optimize $\mathbf{B}$ in a LoRA module during training, which is much more efficient. Besides, we can store the random seed that generates the random projection of $\mathbf{A}$, requiring much less memory.
	We empirically verify that LoRA only introduces a quite small amount of extra trainable parameters that are less than 1\% of the original parameters.
	
	\subsection{Reliability Consolidation with LoRA Arithmetic}
	Let $\theta_{\rm pre} \in \mathbb{R}^{M}$ be the parameters of a pre-trained model. To deal with failure detection in the wild, we freeze $\theta_{\rm pre}$ and learn an additional LoRA module with a loss function related to a specific emerged failure at phase $t$ (e.g., OE loss for semantic shifts).
	Let $\theta_{{\rm LoRA}, t-1} \in \mathbb{R}^{m}$ be the weights of the LoRA before fine-tuning, $\theta_{{\rm LoRA}, t} \in \mathbb{R}^{m}$ be the corresponding weights after fine-tuning and $m \ll M$. 
	The LoRA vector $\tau_{{\rm LoRA}, t}$ is given by the element-wise difference between \{$\theta_{\rm pre}$, $\theta_{{\rm LoRA}, t}$\} and \{$\theta_{\rm pre}$, $\theta_{{\rm LoRA}, t-1}$\} as follow:
	\begin{equation}
		\begin{aligned}
			\tau_{{\rm LoRA}, t} = \{\theta_{\rm pre}, \theta_{{\rm LoRA}, t}\} - \{\theta_{\rm pre},\theta_{{\rm LoRA}, t-1}\} = \theta_{{\rm LoRA}, t} - \theta_{{\rm LoRA}, t-1}.
		\end{aligned}
	\end{equation}
	The intuition behind LoRA vector is to encapsulate crucial
	directions in which the model’s parameters move when learning with a loss function \cite{ilharco2022editing} dealing with a specific failure source. 
    
	As illustrated in Fig.~\ref{figure-3} (Right), after fine-tuning each LoRA module with its respective learning objective, we can perform reliability enhancement or reduction easily and flexibly via element-wise
	addition or negation with a scaling term $\alpha \in [0,1]$ as follows: \textbf{LoRA addition:} The sum of the LoRA vectors $\tau = \sum_{t} \alpha_{t}\tau_{{\rm LoRA}, t}$ is added to a pre-trained model $\theta_{\rm pre}$ to produce a model that performs failure detection on different failure sources. In our cases, we focus on model reliability under both covariate and semantic shifts, and we can get a model $\{\theta_{\rm pre}, \tau\}$ with unified failure detection ability by merging the two LoRA vectors trained using AugMix and OE easily, in which
		\begin{equation}\label{add}
			\begin{aligned}
				\tau = (1-\alpha) \cdot \tau_{{\rm LoRA, cov}} + \alpha \cdot \tau_{{\rm LoRA, sem}}.
			\end{aligned}
		\end{equation}
	\textbf{LoRA negation:} We can reduce the ability of rejecting specific failure
		while retaining performance in other cases by subtracting the LoRA vector from the given LoRA-augmented model. For example, we can get a model $\{\theta_{\rm pre}, \tau\}$, whose OOD detection ability is weaken with $\tau = - \alpha \cdot \tau_{{\rm LoRA, sem}}$.
    
	The LoRA arithmetic is simple and effective to address the challenging unified failure detection. Specifically, in our case, we get LoRA vectors regarding covariate and semantic shifts via learning objectives presented in Eq. (\ref{augmix}) and Eq. (\ref{oe}), respectively. Then we perform LoRA addition to consolidate those two aspects of reliability, having the following advantages: \textbf{(1)} \textbf{\emph{Flexible.}} the scaling term $\alpha$ provides the possibility and flexibility to control the strength of reliability edition, easily adjusting the trade-off without full retraining. \textbf{(2)} \textbf{\emph{Efficient.}} When facing new failure cases, we only fine-tune the LoRA, which is lightweight and computationally efficient with minimal latency compared with full training. \textbf{({3})} \textbf{\emph{Recoverable.}} We can easily recover the model to the default setting without losing the original knowledge by removing the LoRA module. 
	
	\section{Experiments}
	\textbf{Datasets and implementation.} Following the common setup in literature, 
	we assume that the real distribution of OOD data remains unknown during training. 
	For covariate-shifted data, we use CIFAR-10/100-C \cite{hendrycks2018benchmarking} consists of 15 diverse corruption types; for semantic-shifted data, we use natural image datasets including {\ttfamily SVHN}, {\ttfamily Textures},
	{\ttfamily Places}, {\ttfamily LSUN-Crop}, {\ttfamily LSUN-Resize}, and {\ttfamily iSUN}. \textbf{\emph{To focus on the failure detection ability on distribution shifts}}, we first evaluate the performance with a mixture of covariate-shifted and semantic-shifted data at the inference stage and generally keep equal numbers of misclassified {covariate}-shifted data $\mathcal{P}^{\rm covariate}_{\rm out}(y \neq h(\mathbf{x}))$ and semantic OOD data $\mathcal{P}^{\rm semantic}_{\rm out}$, which are two kinds of failure sources we want to reject. \textbf{\emph{Then, we provide the unified failure detection}} results evaluated on both clean ID and distribution-shifted data.
	The standard pre-trained model is fine-tuned for 10 epochs using a cosine learning rate with an initial learning rate of 0.001 to acquire different aspects of reliability. For LoRA, we simply set $r=4$. 
	
	\begin{table*}[t]
		\caption{Failure detection performance under mixture of covariate and semantic shifts on CIFAR-10 with ResNet-18. Methods with $^{\ast}$ train from scratch, methods with $^{+}$ fully fine-tune the pretrained model,  while others only fine-tune the LoRA.}
		\vskip -0.07in
		\label{table-1}
		\setlength\tabcolsep{5pt}
		\centering
		\renewcommand{\arraystretch}{1.05}
		\scalebox{0.82}{
			\begin{tabular}{l|cccc|cccc|cccc}
				\toprule[1.5pt]
				\multicolumn{1}{l}{\multirow{2}{*}{Method}} & \multicolumn{4}{c}{Severity-1} & \multicolumn{4}{c}{Severity-2} & \multicolumn{4}{c}{Severity-3}\\
				\cmidrule(lr){2-5} \cmidrule(lr){6-9} \cmidrule(lr){10-13}
				&\textbf{AURC} &FPR95&AUC  &F-AUC  &\textbf{AURC} &FPR95&AUC  &F-AUC &\textbf{AURC} &FPR95&AUC  &F-AUC \\
				\midrule[1.2pt]
				CE$^{\ast}$ & 56.04 & 37.53 & 89.49  & 87.41 & 89.05 & 44.23 & 86.93  & 84.19 & 122.75 &  48.93 & 84.63  & 81.17 \\
				RegMixUp$^{\ast}$ &57.05 & 50.60 & 88.56  & 86.86 & 89.13 & 55.62 & 86.30  & 83.89 & 124.50 & 58.52 & 84.05  &80.99\\
				CRL$^{\ast}$ & 50.25 & 32.06 & 90.35  & 88.12 & 81.56 & 38.47 & 87.98  & 84.93 & 115.06 & 44.28 & 85.58  & 81.59 \\
				LogitNorm$^{\ast}$ & 48.15 & 32.56 & 91.92  & 90.88 & 76.79 & 38.44 & 90.01  & 88.53 & 107.45 & 43.63 & 88.25  & 86.25 \\
				OE$^{\ast}$ & 51.24 & 33.30 & 91.73  & 90.37 & 82.77 & 39.12 & 89.76  & 87.51 & 120.52 & 44.37 & 87.46  & 84.20 \\
				{OpenMix$^{\ast}$} &{29.46} &{28.13} &{92.45} &{91.08} &{46.90} &{32.61} &{90.95} &{89.12} &{66.86} &{36.94} &{89.18} &{86.83}\\
				{SURE$^{\ast}$} &{31.25} &{27.39} &{92.67} &{91.26} &{48.13} &{31.19} &{91.02} &{89.51} &{68.31} &{36.60} &{89.55} &{86.27}\\
				{RCL$^{+}$} &{58.01} &{35.92} &{89.53} &{87.47} &{93.19} &{43.11} &{86.95} &{84.17} &{132.03} &{48.36} &{84.57} &{81.12}\\
				{SCONE$^{+}$} &{44.01} &{26.99} &{93.13} &{92.08} &{71.28} &{32.75} &{91.17} &{89.37} &{104.06} &{38.91} &{88.83} &{86.15}\\
				\midrule
				RegMixUp & 62.17 & 53.63 & 87.97  & 86.11 & 94.89 & 57.20 & 85.73  & 83.23 & 130.78 & 61.87 & 83.37  & 80.32 \\
				CRL & 56.28 & 38.40 & 89.36  & 87.37 & 88.87 & 44.57 & 86.97  & 84.27 & 125.28 & 50.16 & 84.45  & 81.09 \\
				LogitNorm & 59.05 & 36.71 & 89.41  & 87.22 & 94.15 & 42.89 & 86.84  & 83.85 & 133.41 & 48.76 & 84.26  & 80.62 \\
				OE &43.63 & 26.01 & 93.51  & 92.47 & 70.04 & 30.86 & 92.56  & 90.69 & 100.99 & 36.04 & 90.91  & 87.96\\
				AugMix &36.62 & 36.09 & 90.72  & 89.16 & 51.73 & 39.06 & 89.61  & 87.71 & 67.82  & 41.97 & 88.51  & 86.20 \\
				MaxLogit &40.58 & 45.39 & 89.30  & 88.08 & 55.32 & 46.52 & 88.57  & 87.06 & 71.02  & 48.31 & 87.74  & 85.90 \\
				Energy &43.50 & 49.19 & 88.03  & 86.84 & 58.75 & 50.10 & 87.33  & 85.94 & 74.90  & 51.51 & 86.48  & 84.86 \\
				KNN &43.87 & 42.53 & 87.20  & 86.00 & 61.94 & 45.98 & 85.71  & 84.27 & 82.10  & 48.63 & 84.02  & 82.35 \\
				{FS-KNN} &{47.54} &{55.75} &{87.29} &{86.01} &{62.36} &{55.35} &{86.79} &{85.30} &{80.36} &{57.86} &{85.52} &{83.74} \\
				NNGuide &51.77 & 63.95 & 85.25  & 84.05 & 68.17 & 63.28 & 84.65  & 83.32 & 84.80  & 63.49 & 84.01  & 82.40 \\
				Relation &58.98 & 59.52 & 80.85  & 79.74 & 77.10 & 60.19 & 80.17  & 78.85 & 97.12  & 61.66 & 79.04  & 77.57 \\
				GEN &41.30 	&46.00 	&88.78 	&87.62 	&56.34 	&47.31 	&88.00 	&86.61 	&72.20 	&48.94 	&87.18 	&85.48 
				\\
				ASH &41.10 & 45.87 & 89.11  & 87.89 & 55.97 & 47.28 & 88.31  & 86.83 & 72.08  & 49.31 & 87.44  & 85.66 \\
				\rowcolor{mygray}
				TrustLoRA &\textbf{28.68} & \textbf{23.80} & \textbf{93.67}  &\textbf{92.53} & \textbf{41.64} &\textbf{27.43} & \textbf{92.67}  &\textbf{91.18} &\textbf{56.64}  &\textbf{30.62} &\textbf{91.55}  &\textbf{89.65}\\
				\bottomrule[1.5pt]
		\end{tabular}}
		\vskip -0.05in
	\end{table*}

    \begin{table*}[t]
		\caption{Failure detection under mixture of covariate and semantic shifts on CIFAR-100.}
		\vskip -0.07in
		\label{table-2}
		\setlength\tabcolsep{5pt}
		\centering
		\renewcommand{\arraystretch}{1.05}
		\scalebox{0.82}{
			\begin{tabular}{l|cccc|cccc|cccc}
				\toprule[1.5pt]
				\multicolumn{1}{l}{\multirow{2}{*}{Method}} & \multicolumn{4}{c}{Severity-1} & \multicolumn{4}{c}{Severity-2} & \multicolumn{4}{c}{Severity-3}\\
				\cmidrule(lr){2-5} \cmidrule(lr){6-9} \cmidrule(lr){10-13}
				&\textbf{AURC} &FPR95&AUC  &F-AUC  &\textbf{AURC} &FPR95&AUC  &F-AUC &\textbf{AURC} &FPR95&AUC  &F-AUC \\
				\midrule[1.2pt]
				CE$^{\ast}$ &163.66 & 54.88 & 82.95 & 76.33 & 218.26 & 61.90 & 79.53 & 71.53 & 279.10 & 67.12 & 76.65 & 68.06 \\
				RegMixUp$^{\ast}$ &154.87 & 54.34 & 83.12 & 76.72 & 198.08 & 60.46 & 80.25 & 72.64 & 249.93 & 65.33 & 77.43 & 69.25 \\
				CRL$^{\ast}$ &152.20 & 53.03 & 83.94 & 77.67 & 195.12 & 59.43 & 81.07 & 73.37 & 244.45 & 64.10 & 78.74 & 70.00 \\
				LogitNorm$^{\ast}$ &166.34 & 56.51 & 82.91 & 77.24 & 217.56 & 62.15 & 80.21 & 73.37 & 273.87 & 66.38 & 77.78 & 70.37 \\
				OE$^{\ast}$ &149.02 & 45.48 & 87.09 & 83.17 & 194.97 & 51.69 & 85.26 & 80.21 & 246.45 & 56.24 & 83.31 & 77.18 \\
				{OpenMix$^{\ast}$} &{134.18} &{46.46} &{86.57} &{81.84} &{164.32} &{51.49} &{84.47} &{78.78} &{203.87} &{56.18} &{82.29} &{76.10}\\
				{SURE$^{\ast}$} &{137.62} &{48.35} &{86.04} &{82.26} &{172.45} &{51.76} &{83.95} &{78.41} &{205.71} &{56.62} &{82.18} &{76.54}\\
				{RCL$^{+}$} &{155.85} &{52.89} &{84.29} &{78.73} &{202.39} &{59.20} &{81.61} &{74.88} &{256.47} &{64.10} &{79.11} &{71.83} \\
				{SCONE$^{+}$} &{148.50} &{47.65} &{86.50} &{81.69} &{201.73} &{53.92} &{83.57} &{77.37} &{264.59} &{59.29} &{80.74} &{74.03}\\
				\midrule
				RegMixUp &155.58 & 53.51 & 83.81 & 77.99 & 203.97 & 59.66 & 80.70 & 73.83 & 261.76 & 64.50 & 77.96 & 70.71 \\
				CRL &153.48 & 52.10 & 84.27 & 78.57 & 204.13 & 58.43 & 81.24 & 74.25 & 262.74 & 63.65 & 78.31 & 71.19 \\
				LogitNorm &155.08 & 52.55 & 84.14 & 78.38 & 207.23 & 58.92 & 81.07 & 74.10 & 267.35 & 63.96 & 78.16 & 71.06 \\
				OE  &147.40 & 46.57 & 87.21 & 83.04 & 197.74 & 52.09 & 84.73 & 79.32 & 259.43 & 57.08 & 82.12 & 76.15 \\
				AugMix &141.23 & 51.24 & 84.69 & 79.69 & 158.07 & 54.44 & 83.47 & 77.63 & 177.71 & 57.12 & 82.26 & 75.69 \\
				MaxLogit &150.37 & 57.84 & 83.33 & 78.72 & 166.31 & 59.89 & 82.35 & 76.96 & 185.83 & 61.91 & 81.20 & 75.10 \\
				Energy &158.37 & 60.62 & 81.52 & 77.58 & 175.02 & 62.48 & 80.47 & 75.90 & 194.12 & 64.27 & 79.40 & 74.28 \\
				KNN &168.58 & 66.07 & 80.55 & 77.54 & 184.77 & 67.07 & 79.41 & 75.79 & 204.44 & 68.17 & 78.22 & 74.09 \\
				{FS-KNN} &{143.83} &{56.08} &{84.71} &{81.69} &{167.36} &{58.13} &{83.03} &{79.74} &{183.17} &{59.21} &{82.26} &{77.12} \\
				NNGuide &180.32 & 66.12 & 76.27 & 72.23 & 197.55 & 67.51 & 75.21 & 70.47 & 215.89 & 68.58 & 74.39 & 69.10 \\
				Relation &171.64 & 67.49 & 80.28 & 76.20 & 186.00 & 67.95 & 79.49 & 74.63 & 204.07 & 68.98 & 78.58 & 73.07 \\
				GEN &157.47 	&60.25 	&81.68 	&77.69 	&174.32 	&62.19 	&80.57 	&75.99 	&193.63 	&64.04 	&79.50 	&74.32 
				\\
				ASH &150.38 & 57.78 & 83.22 & 78.77 & 167.33 & 59.93 & 82.09 & 76.90 & 185.92 & 61.75 & 81.05 & 75.13 \\
				\rowcolor{mygray}
				TrustLoRA &\textbf{129.64} & \textbf{46.46} & \textbf{87.29} & \textbf{83.73} & \textbf{149.14} & \textbf{50.12} & \textbf{85.77} & \textbf{81.40} & \textbf{172.35} & \textbf{53.32} & \textbf{84.41} & \textbf{79.28} \\
				\bottomrule[1.5pt]
		\end{tabular}}
		\vskip -0.1in
	\end{table*}
	
	\textbf{Metrics and compared methods.} We leverage Area Under Risk-Coverage (AURC, \text{\textperthousand}) \cite{GeifmanE17}, FPR95 (\%) and the area under the ROC curve (AUC, \%) \cite{hendrycks2016baseline} to evaluate the performance of failure detection. Besides, we also introduce the F-AUC (\%) i.e., $(2 \times {\rm AUC_{cov}} \times  {\rm AUC_{sem}}) / ({\rm AUC_{cov}} +  {\rm AUC_{sem}})$, where ${\rm AUC_{cov}}$ denotes the AUC value of separating correct and incorrect covariate-shifted data and ${\rm AUC_{cov}}$ denotes the AUC value of separating covariate-shifted and semantic-shifted data. 
	We compare with various methods including CE (MSP) \cite{hendrycks2016baseline}, RegMixUp \cite{pinto2022RegMixup}, CRL \cite{moon2020confidence}, LogitNorm \cite{wei2022mitigating}, OE \cite{hendrycks2018deep}, {OpenMix \cite{zhu2023openmix}, SURE \cite{li2024sure}, RCL \cite{zhu2024rcl},} AugMix \cite{hendrycks2019augmix}, MaxLogit \cite{hendrycks2019anomalyseg}, Energy \cite{liu2020energy}, KNN \cite{sun2022out}, {FS-KNN \cite{cen2023devil},} NNGuide \cite{park2023nearest}, Relation \cite{kim2023neural}, GEN \cite{liu2023gen} and ASH \cite{djurisic2022extremely}. For training-time methods, we report the results of both training from
	scratch and LoRA fine-tuning. Score-based methods are applied to LoRA-augmented model tuning with AugMix. TrustLoRA leverages MSP score.

	\subsection{Results and Discussion}

    \begin{wrapfigure}{r}{9cm}
		\vskip -0.2in
		\includegraphics[width=9cm]{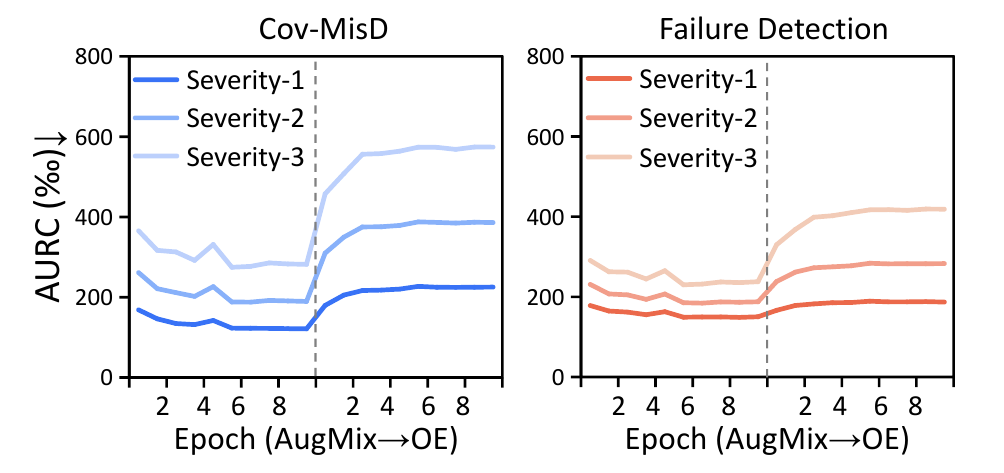}
		\vskip -0.1in
		\caption{Change of rejection ability when fine-tuning the pre-trained ResNet-18 on CIFAR-100.}
		\vskip -0.1in
		\label{figure-4}
	\end{wrapfigure}
	To fully reflect the failure detection performance under both covariate and semantic shifts, we combine each of 15 corruptions under three different severity with six semantic OOD sets, resulting in 90 wild data mixtures in total. We report the average performance on those 90 evaluations. 

    \textbf{Trade-off between the two failure detection tasks.} Fig.~\ref{figure-4} shows the performance change when fine-tuning the pre-trained model.
	Cov-MisD denotes the ability to reject misclassified covariate-shifted (e.g., {\ttfamily Gaussian noise}) examples. Failure detection denotes the ability to reject both misclassified covariate-shifted data and semantic-shifted data jointly. We observe that when fine-tuning with OE (after the dotted line) to acquire OOD detection ability, it is harder to detect misclassified corrupted samples, e.g., AURC ($\downarrow$) of Cov-MisD increases dramatically, and the unified failure detection performance drops.

	\textbf{Our method achieves strong performance.} The main results in Table \ref{table-1} and \ref{table-2} verify that TrustLoRA establishes overall
	strong performance, especially on \textbf{\emph{AURC $\downarrow$, which has been considered as the most important metric for failure detection evaluation}} \cite{jaeger2022call, moon2020confidence}.
	In particular, we consider two groups of baselines: training the model from scratch (denotes with $^{\ast}$) and fine-tuning the pre-trained model. 
	We highlight a few observations: (1) TrustLoRA outperforms strong training methods like LogitNorm \cite{wei2022mitigating}, CRL \cite{moon2020confidence} and RegMixUp \cite{pinto2022RegMixup} in both training from scratch and fine-tuning scenarios. (2) TrustLoRA outperforms
	competitive post-hoc OOD detection methods, which are applied to the same model fine-tuned with AugMix and hence they have the same classification accuracy. (3) The proposed framework excels in detecting both misclassified covariate-shifted and semantic-shifted data, achieving the best performance.
	
	\begin{figure*}[t]
		\begin{center}
	\centerline{\includegraphics[width=1.03\textwidth]{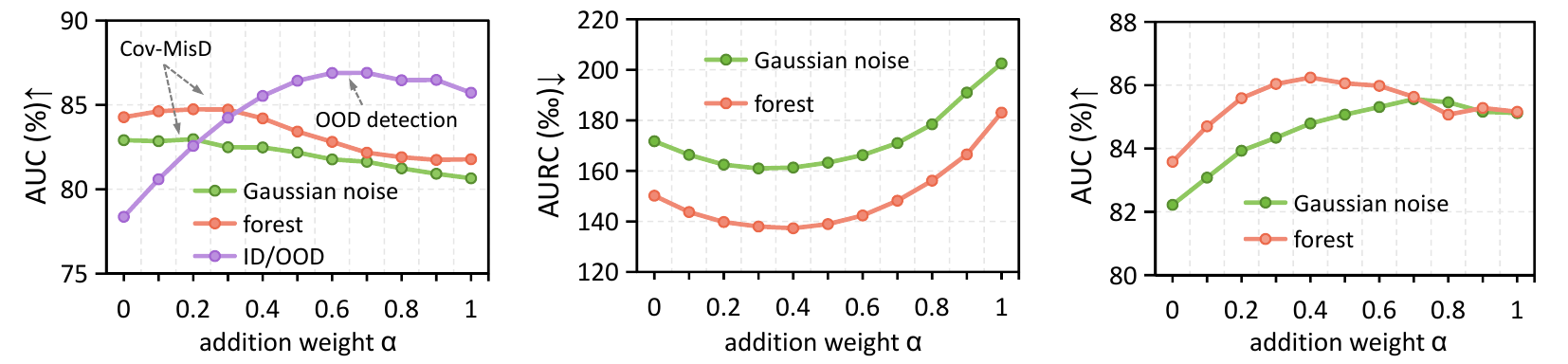}}
			\vskip -0.1 in
			\caption{Flexibility of controlling the strength of reliability edition on CIFAR-100.}
			\label{figure-5}
		\end{center}
		\vskip -0.1 in
	\end{figure*}

	\textbf{Flexibility of controlling the strength of reliability edition.} We separate reliability knowledge with LoRAs and merge them to get a unified failure detector. One of the primary advantages of our method is to control the strength of each kind of reliability flexibly based on end-user preference without training the model again or affecting the original model. In Fig.~\ref{figure-5} (Left), we show that the scaling $\alpha$ in Eq.~(\ref{add}) can easily control the preference between MisD under covariate shifts and OOD detection. In Fig.~\ref{figure-5} (Middle and Right), we observe that an overall strong unified failure detection performance can be achieved with $\alpha \in [0.4,0.6]$, and we simply set $\alpha =0.5$ for all experiments.

    \begin{table}[t]
		\caption{Experimental results on ImageNet.}
		\vskip -0.07in
		\label{table-4}
		\setlength\tabcolsep{17pt}
		\centering
		\renewcommand{\arraystretch}{1.05}
		\scalebox{0.85}{
			\begin{tabular}{lcccc}
				\toprule[1.5pt]
				{\multirow{2}{*}{Method}}                & \multicolumn{2}{c}{ImageNet-200}  & \multicolumn{2}{c}{ImageNet-500}  \\
				\cmidrule(lr){2-3} \cmidrule(lr){4-5}
				& \textbf{AURC}   & AUC            & \textbf{AURC}   & \textbf{AUC}   \\
				\midrule
				CE$^{\ast}$             & 188.37          & 92.55          & 268.42          & 89.08          \\
				MaxLogit$^{\ast}$             & 198.44          & 90.11          & 286.90           & 85.03          \\
				Energy$^{\ast}$               & 203.02          & 89.28          & 295.91          & 83.71          \\
				AugMix (LoRA)         & 166.53          & 93.25          & 236.25          & 89.95          \\
				OE (LoRA)             & 180.18          & 92.78          & 265.11          & 89.50          \\
				MaxLogit (AugMixLoRA) & 187.54          & 91.97          & 259.65          & 87.82          \\
				Energy (AugMixLoRA)   & 197.60          & 91.63          & 266.37          & 86.65          \\
				\rowcolor{mygray}
				TrustLoRA             & \textbf{159.67} & \textbf{93.91} & \textbf{229.92} & \textbf{90.15}\\
				\bottomrule[1.5pt]
			\end{tabular}
		}
		\vskip -0.05 in
	\end{table}

    \begin{wrapfigure}{r}{9cm}
		\vskip -0.2in
		\includegraphics[width=9cm]{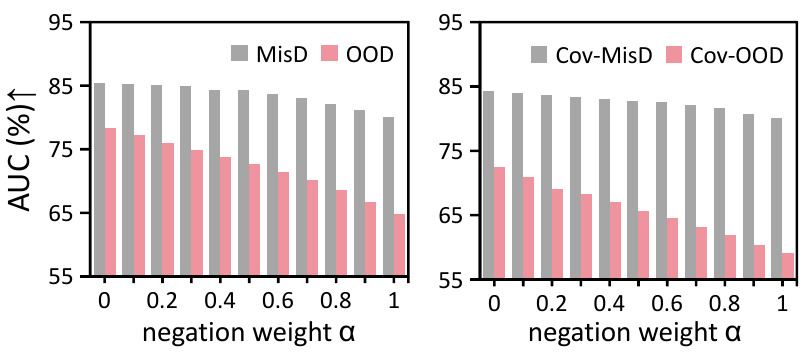}
		\vskip -0.1in
		\caption{Accurate forgetting of OOD detection ability while keeping the MisD ability on clean ID (Left) and covariate-shifted data (Right).}
		\vskip -0.2in
		\label{figure-6}
	\end{wrapfigure}
    \textbf{Large scale experiments on ImageNet.} We provide additional large-scale results
	on the ImageNet-200/500 benchmark with ResNet-50. The classes were randomly sampled from 1K, and we also sampled another set of classes (with equal numbers) as outliers for OE. At inference stage,
    we use a mixture of covariate and semantic OOD data. Specifically, for semantic shifts, we use the fixed ImageNet OOD dataset proposed in \cite{bitterwolf2023or}, which includes truly OOD versions of 11 popular OOD datasets with in total of 2715 OOD samples; for covariate shifts, we use the corruption type {\ttfamily Frost} with severity-1. Results in Table~\ref{table-3} suggest that our method yields strong failure detection performance compared with competitive baselines.

	\textbf{Selective reliability forgetting with LoRA negation.} Besides LoRA addition for unified failure detection, here we explore accurate reliability forgetting. We apply the OOD detection vector  $\tau = - \alpha \cdot \tau_{{\rm LoRA, sem}}$ (learned with OE) to a given model (e.g., LoRA tuning with AugMix). The experiments on CIFAR-100 in Fig.~\ref{figure-6} show that we can enable the model to forget OOD detection ability, while with little deterioration of MisD ability on clean ID (Left) and covariate-shifted data (Right). This demonstrates that our method can enable flexible reliability knowledge edition.

	\begin{table}[t]
		\caption{Comparison of unified failure detection ability evaluated on both clean ID and distribution-shifted data.}
		\vskip -0.07in
		\label{table-66}
		\setlength\tabcolsep{8pt}
		\centering
		\renewcommand{\arraystretch}{1.05}
		\scalebox{0.85}{
			\begin{tabular}{l|cccc|cccc}
				\toprule[1.5pt]
				\multicolumn{1}{l}{\multirow{2}{*}{{Method}}} & \multicolumn{4}{c}{{CIFAR-10}} & \multicolumn{4}{c}{{CIFAR-100}} \\
				\cmidrule(lr){2-5} \cmidrule(lr){6-9}
				&\textbf{{AURC}} &{FPR95} &{AUC}  &{F-AUC}  &\textbf{{AURC}} &{FPR95} &{AUC}  &{F-AUC} \\
				\midrule[1.2pt]
				{CE} &{32.62} &{31.56} &{91.59} &{89.98} &{134.19} &{48.70} &{85.47} &{79.71} \\
				{RegMixUp} &{34.82} &{44.92} &{90.56} &{89.27} &{133.88} &{49.25} &{84.96} &{79.35} \\
				{CRL} &{30.37} &{26.24} &{92.30} &{90.54} &{129.75} &{47.73} &{85.97} &{80.47} \\
				{LogitNorm} &{27.11} &{26.10} &{93.76} &{92.97} &{138.73} &{51.38} &{85.22} &{80.20} \\
				\midrule
				{AugMix} &{27.47} &{31.75} &{91.85} &{90.45} &{133.73} &{49.50} &{85.34} &{80.70} \\
				{MaxLogit} &{32.53} &{46.17} &{89.95} &{88.86} &{142.31} &{56.11} &{84.00} &{79.74} \\
				{Energy} &{35.48} &{50.48} &{88.52} &{87.49} &{149.95} &{58.96} &{82.23} &{78.55} \\
				{GEN} &{33.33} &{46.73} &{89.31} &{88.31} &{149.52} &{58.66} &{82.30} &{78.59} \\
				{ASH} &{33.09} &{46.95} &{89.67} &{88.61} &{142.67} &{56.05} &{83.87} &{79.75} \\
				\rowcolor{mygray}
				{TrustLoRA} &\textbf{{20.86}} & \textbf{{20.86}} & \textbf{{94.42}} & \textbf{{93.47}} & \textbf{{121.90}} & \textbf{{44.44}} & \textbf{{87.75}} & \textbf{{84.35}} \\
				\bottomrule[1.5pt]
		\end{tabular}}
	\end{table}
	
	\begin{figure}[t]
		\begin{center}
	\centerline{\includegraphics[width=0.75\textwidth]{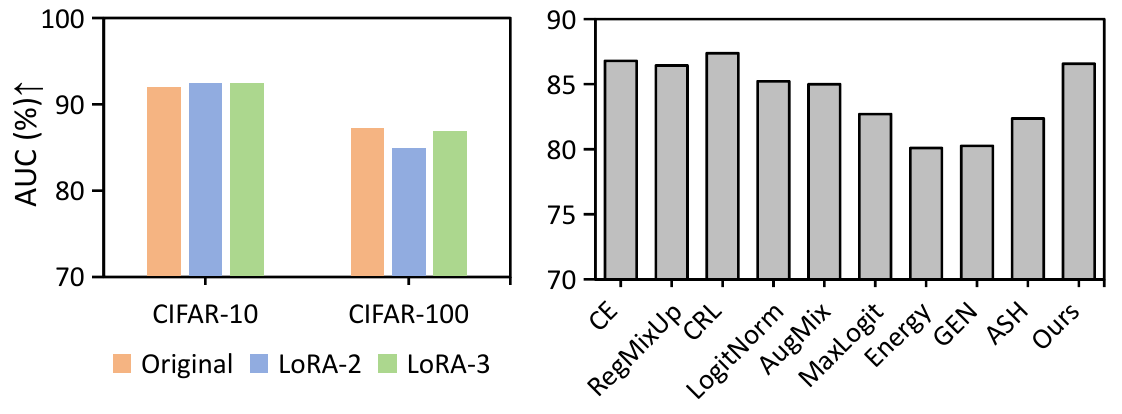}}
			\vskip -0.1 in
			\caption{MisD ability on clean ID.}
			\label{figure-99}
		\end{center}
		\vskip -0.2 in
	\end{figure}

	\begin{table*}[t]
		\caption{Failure detection performance on CIFAR-100 with ViT.}
		\vskip -0.07in
		\label{table-3}
		\setlength\tabcolsep{5pt}
		\centering
		\renewcommand{\arraystretch}{1.05}
		\scalebox{0.85}{
			\begin{tabular}{l|cccc|cccc|cccl}
				\toprule[1.5pt]
				\multicolumn{1}{l}{\multirow{2}{*}{Method}} & \multicolumn{4}{c}{Severity-1} & \multicolumn{4}{c}{Severity-2} & \multicolumn{4}{c}{Severity-3}\\
				\cmidrule(lr){2-5} \cmidrule(lr){6-9} \cmidrule(lr){10-13}
				&\textbf{AURC} &FPR95&AUC  &F-AUC  &\textbf{AURC} &FPR95&AUC  &F-AUC &\textbf{AURC} &FPR95&AUC  &F-AUC \\
				\midrule[1.2pt]
				Full-FT &49.17 & 33.80 & 91.07 & 89.09 & 69.78 & 37.67 & 89.77 & 87.08 & 92.46 & 41.43 & 88.39 & 84.81 \\
				Linear &98.88 & 37.34 & 90.03 & 87.23 & 116.38 & 40.22 & 88.94 & 85.33 & 134.55 & 43.36 & 87.70 & 83.20 \\
				\midrule
				CE &51.68 & 33.77 & 91.30 & 89.61 & 73.33 & 36.91 & 90.14 & 87.88 & 93.37 & 40.73 & 88.87 & 85.95 \\
				RegMixUp & 50.54 & 37.70 & 90.83 & 89.21 & 74.42 & 44.21 & 88.97 & 86.78 & 97.86 & 49.19 & 87.35 & 84.41 \\
				CRL & 54.48 & 34.01 & 91.23 & 89.44 & 76.32 & 37.63 & 90.01 & 87.65 & 95.83 & 40.86 & 88.74 & 85.70 \\
				LogitNorm &59.66 & 49.24 & 88.70 & 87.39 & 81.65 & 50.18 & 88.16 & 86.44 & 101.24 & 51.96 & 87.25 & 85.03 \\
				\midrule
				AugMix &46.13 & 34.27 & 91.12 & 89.37 & 65.15 & 37.65 & 90.06 & 87.75 & 85.36 & 41.12 & 88.83 & 85.89 \\
				MaxLogit & 52.22 & 43.66 & 89.35 & 88.08 & 70.65 & 45.35 & 88.76 & 87.11 & 89.95 & 47.21 & 87.93 & 85.79 \\
				Energy & 55.52 & 48.50 & 88.24 & 87.00 & 74.24 & 49.68 & 87.76 & 86.18 & 93.74 & 51.18 & 87.02 & 85.07 \\
				KNN & 58.91 & 53.46 & 86.82 & 85.46 & 78.90 & 54.02 & 86.11 & 84.51 & 99.20 & 54.78 & 85.32 & 83.38 \\
				NNGuide & 	53.86 & 46.97 & 88.53 & 87.28 & 72.53 & 48.36 & 87.96 & 86.40 & 92.06 & 50.20 & 87.17 & 85.23 \\
				GEN & 54.15 & 46.31 & 88.59 & 87.36 & 73.01 & 47.82 & 88.00 & 86.48 & 92.40 & 49.49 & 87.25 & 85.28 \\
				\rowcolor{mygray}
				TrustLoRA &\textbf{43.40} & \textbf{31.25} & \textbf{92.13} & \textbf{90.78} & \textbf{63.01} & \textbf{34.08} & \textbf{91.22} & \textbf{89.36} & \textbf{83.37} & \textbf{37.91} & \textbf{89.90} & \textbf{87.56} \\
				\bottomrule[1.5pt]
		\end{tabular}}
	\end{table*}
	
	\textbf{Unified failure detection on clean ID and distribution-shifted data.} In above experiments, to \emph{clearly reflect the failure detection ability under covariate and semantic shits}, we do not include clean ID data at inference time. With integrated LoRAs of covariate and semantic shifts, the MisD performance on the original clean set can be well preserved on CIFAR-10 while suffering from a slight drop on CIFAR-100. As shown in Fig.~\ref{figure-99} (Left), TrustLoRA can further recover and integrate MisD knowledge on clean set by merging an additional LoRA fine-tuned with flat minima loss \cite{zhu2023revisiting}, and we denote the model ``LoRA-3''. Fig.\ref{figure-99} (Right) compares the MisD on clean ID data, where our method successfully achieves comparable MisD performance with the original model, and outperforms other methods. 
	Table \ref{table-66} further reports the results on full spectrum of test set including clean ID, covariate and semantic OOD data. 
	As can be observed, our method still achieves strong performance and outperforms other methods.
	
	\textbf{Experiments with ViT.}
	We also conduct experiments on pre-trained ViT backbone (ViT-B16) \cite{dosovitskiy2020image}, and perform full fine-tuning, linear prob and LoRA tuning. We fine-tuned the model for 10 epochs using cosine learning rate with the initial learning rate
	of 0.03. We set the momentum to be 0.9 and the weight
	decay to 0. Despite the strong performance of pre-trained ViT-B16, results in Table~\ref{table-3} reveal that our method yields notable improvement, especially on AURC.

	\textbf{TrustLoRA outperforms the multi-task learning.} We further compare our method with more baselines: (1) SIRC \cite{xia2024augmenting} and FMFP \cite{zhu2023revisiting} (FlatLoRA in our comparison). (2) Multi-task tuning with combined OE and AugMix learning objectives. The results in Table~\ref{table-5} verify that our method outperforms SIRC and FlatLoRA consistently. In particular, our LoRA arithmetic outperforms the multi-task learning, i.e., AugMix+OE (Full FT) in Table~\ref{table-5}. \textbf{\emph{Intuitively}}, this is because when optimizing both two objectives in a multi-task learning (MTL) manner, there exist remarkable conflicts between pulling covariate-shifted samples close to class centers while pushing semantic-shifted samples away from class centers since those two types of shifted samples are often overlapping. \textbf{\emph{Theoretically}}, the Bayes-optimal reject rule for MisD is based on maximum class-posterior probability ${\max}_{y \in \mathcal{Y}}\mathbb{P}(y|\textbf{x})$, while OOD detection rejects samples with small density ratio $p(\textbf{x}|\text{in}) / p(\textbf{x}|\text{out})$ \cite{zhu2023revisiting}. However, to separate samples from known classes and unknown semantic-shifted unknown classes, binary discrimination would compress the confidence distribution of correct and incorrect covariate-shifted samples. As a result, MTL of the two objectives would be unstable. Differently, our proposed LoRA arithmetic overcomes the above limitation with reliability knowledge separation and consolidation.
    
	\begin{table}[t]
\centering
		\caption{Comparison with more baselines and multi-task learning on CIFAR-100, severity-1.}
		\vskip -0.07in
		\label{table-5}
			\setlength\tabcolsep{6pt}
		\centering
		\renewcommand{\arraystretch}{1.05}
		\scalebox{0.85}{
			\begin{tabular}{l|cccc|cccc}
				\toprule[1.5pt]
				\multicolumn{1}{l}{\multirow{2}{*}{Method}} & \multicolumn{4}{c}{Severity-1} & \multicolumn{4}{c}{Severity-2}\\
				\cmidrule(lr){2-5} \cmidrule(lr){6-9}
				&\textbf{AURC} &FPR95&AUC  &F-AUC  &\textbf{AURC} &FPR95&AUC  &F-AUC \\
				\midrule[1.2pt]
SIRC$^{\ast}$ (MSP,z1)      & 160.26 & 52.17 & 83.75 & 76.65 & 214.24 & 60.07 & 80.22 & 71.85 \\
FlatLoRA            & 152.94 & 51.97 & 84.40  & 78.74 & 201.94 & 58.51 & 81.43 & 74.46 \\
SIRC (AugMixLoRA)   & 139.86 & 50.53 & 85.50  & 80.04 & 156.63 & 54.03 & 83.86 & 78.25 \\
AugMix (Full FT)    & 133.68 & 49.79 & 85.05 & 80.48 & 146.83 & 52.47 & 83.92 & 78.66 \\
AugMix+OE (Full FT) & 138.92 & 50.26 & 85.40  & 81.44 & \textbf{142.66} & 50.66 & 84.52 & 80.12 \\
\rowcolor{mygray}
TrustLoRA & \textbf{129.64} & \textbf{46.46} & \textbf{87.29} & \textbf{83.73} & 149.14 & \textbf{50.12} & \textbf{85.77} & \textbf{81.40}\\
				\bottomrule[1.5pt]
		\end{tabular}}
        \vskip -0.1in
\end{table}

\begin{table}[t]
\centering
		\caption{ {Performance of TrustLoRA with different rank $r$ on CIFAR-10/100, severity-1.}}
		\vskip -0.07in
		\label{table-99}
			\setlength\tabcolsep{6.5pt}
		\centering
		\renewcommand{\arraystretch}{1.05}
		\scalebox{0.85}{
			\begin{tabular}{l|cccc|cccc}
				\toprule[1.5pt]
				\multicolumn{1}{l}{\multirow{2}{*}{ {Method}}} & \multicolumn{4}{c}{ {CIFAR-10}} & \multicolumn{4}{c}{ {CIFAR-100}}\\
				\cmidrule(lr){2-5} \cmidrule(lr){6-9}
				&\textbf{ {AURC}} & {FPR95} & {AUC}  & {F-AUC}  &\textbf{ {AURC}} & {FPR95} & {AUC}  & {F-AUC} \\
				\midrule[1.2pt]
 {$r=2$}      &{29.31}  &{24.92}  &{93.32}  &{92.24}  &{131.06} &{47.17}  &{87.11}  &{83.38} \\
 {$r=4$}            &{28.68}  &{23.80}  &{93.67}  &{92.53}  &{129.64} &{46.46}  &{87.29}  &{83.73}  \\
 {$r=8$}   &{28.45}  &{23.11}  &{93.60}  &{92.72}  &{128.15} &{46.03}  &{87.40}  &{83.82}  \\
				\bottomrule[1.5pt]
		\end{tabular}}
\end{table}

\textbf{Rank of the LoRA.} A higher rank $r$ in LoRA means a greater number of trainable parameters and might lead to overfitting, while a lower rank $r$ means fewer trainable parameters and might lead to underfitting.  
When fine-tuning a pretrained model, if the dataset is significantly different and more complex, then it’s would be better to use a high rank value (e.g., 64–256). On the other hand, if there doesn’t involve a complex new dataset that the model hasn’t encountered before, lower values of rank (e.g., 4–12) are sufficient. We conduct experiments under different settings of rank in LoRA (Severity-1). As shown in Table \ref{table-99}, our method is robust to the different ranks of LoRA, and we simply set $r=4$ (which is a common choice for LoRA tuning) for all experiments in our main paper. In our case, we aim to learn reliability knowledge about the current task, without introducing a complex new dataset. Therefore, low value of rank like 4 or 2 is sufficient to learn the additional reliability knowledge without underfitting.

\begin{wraptable}{R}{0.6\linewidth}
\centering
\vskip -0.15in
\caption{Comparison of the computational costs.}
\vskip -0.1in
\label{table-6}
\renewcommand{\arraystretch}{1}
\resizebox{\linewidth}{!}{
\begin{tabular}{lccc}
\toprule[1.3pt]
Model     & ResNet-20    & ResNet-18 &    ViT (B16)        \\
\midrule
BaseModel & 0.2871$M$ & 10.91$M$ & 81.89$M$\\
TrustLoRA & \textbf{0.0275$M$} & \textbf{0.25$M$} & \textbf{0.21$M$} \\
\bottomrule[1.3pt]
\end{tabular}}
\vskip -0.1in
\end{wraptable}
 \textbf{Computational cost with random projection based LoRA.} In our method, we propose LoRA with random projection, where only the $\mathbf{B}$ matrix of LoRA is trained. In Table \ref{table-999}, we show that random projection based LoRA achieves similar with that training both $\mathbf{A}$ and $\mathbf{B}$, while needing less computation and memory cost.
Table~\ref{table-6} reports the number of parameters of the base model and all LoRA modules, where our method has much smaller parameters than that of base model. Since we fix the parameters of $\mathbf{A}$ once initialized and only optimize $\mathbf{B}$ in LoRA during the training stage, which is much more efficient than learning the original LoRA. 
	
\begin{table}[t]
\centering
		\caption{Random projection based LoRA \emph{v.s.} the original LoRA on CIFAR-10/100, severity-1.}
		\vskip -0.07in
		\label{table-999}
			\setlength\tabcolsep{6.5pt}
		\centering
		\renewcommand{\arraystretch}{1.05}
		\scalebox{0.85}{
			\begin{tabular}{l|cccc|cccc}
				\toprule[1.5pt]
				\multicolumn{1}{l}{\multirow{2}{*}{ {Method}}} & \multicolumn{4}{c}{ {CIFAR-10}} & \multicolumn{4}{c}{ {CIFAR-100}}\\
				\cmidrule(lr){2-5} \cmidrule(lr){6-9}
				&\textbf{ {AURC}} & {FPR95} & {AUC}  & {F-AUC}  &\textbf{ {AURC}} & {FPR95} & {AUC}  & {F-AUC} \\
				\midrule[1.2pt]
 {B only}           &{28.68}  &{23.80}  &{93.67}  &{92.53}  &{129.64} &{46.46}  &{87.29}  &{83.73}  \\ 
 {A \& B}           &{28.07}  &{24.12}  &{93.82}  &{92.95}  &{127.51} &{45.72}  &{87.17}  &{83.34}  \\
				\bottomrule[1.5pt]
		\end{tabular}}
\end{table}

\begin{table*}[t]
    \caption{ {Robustness to different auxiliary data when acquiring OOD detection ability.}}
    \vskip -0.07in
    \label{table-22}
    \setlength\tabcolsep{5pt}
    \centering
    \renewcommand{\arraystretch}{1.05}
    \scalebox{0.85}{
    \begin{tabular}{lcccccccc}
        \toprule[1.3pt]
        \multicolumn{1}{l}{\multirow{2}{*}{ {Auxiliary Data}}} & \multicolumn{4}{c}{ {Severity-1}} & \multicolumn{4}{c}{ {Severity-2}}\\
        \cmidrule(lr){2-5} \cmidrule(lr){6-9}
        &\textbf{ {AURC}} & {FPR95} & {AUC}  & {F-AUC}  &\textbf{ {AURC}} & {FPR95} & {AUC}  & {F-AUC} \\
        \midrule
        {\ttfamily  {TIN597}} & {130.45} &  {45.26} &  {87.41} &  {83.19} &  {152.50} &  {49.63} &  {85.81} &  {80.83} \\
        {\ttfamily  {RandomImage}} & {129.64} &  {46.46} &  {87.29} &  {83.73} &  {149.14} &  {50.12} &  {85.77} &  {81.40}\\
        \bottomrule[1.3pt]
    \end{tabular}}
\end{table*}
\textbf{Robustness to different auxiliary data.} In this paper, the proposed TrustLoRA acquires OOD detection via OE technique, which requires access to auxiliary outlier data. For CIFAR benchmark, the {\ttfamily RandomImage} is used as auxiliary outliers following existing work. In Table~\ref{table-22} (CIFAR-100), we show that TrustLoRA is robust to other auxiliary outliers like TIN597 \citep{zhang2023openood}.

\begin{wrapfigure}{r}{9cm}
\vskip -0.18in
		\includegraphics[width=9cm]{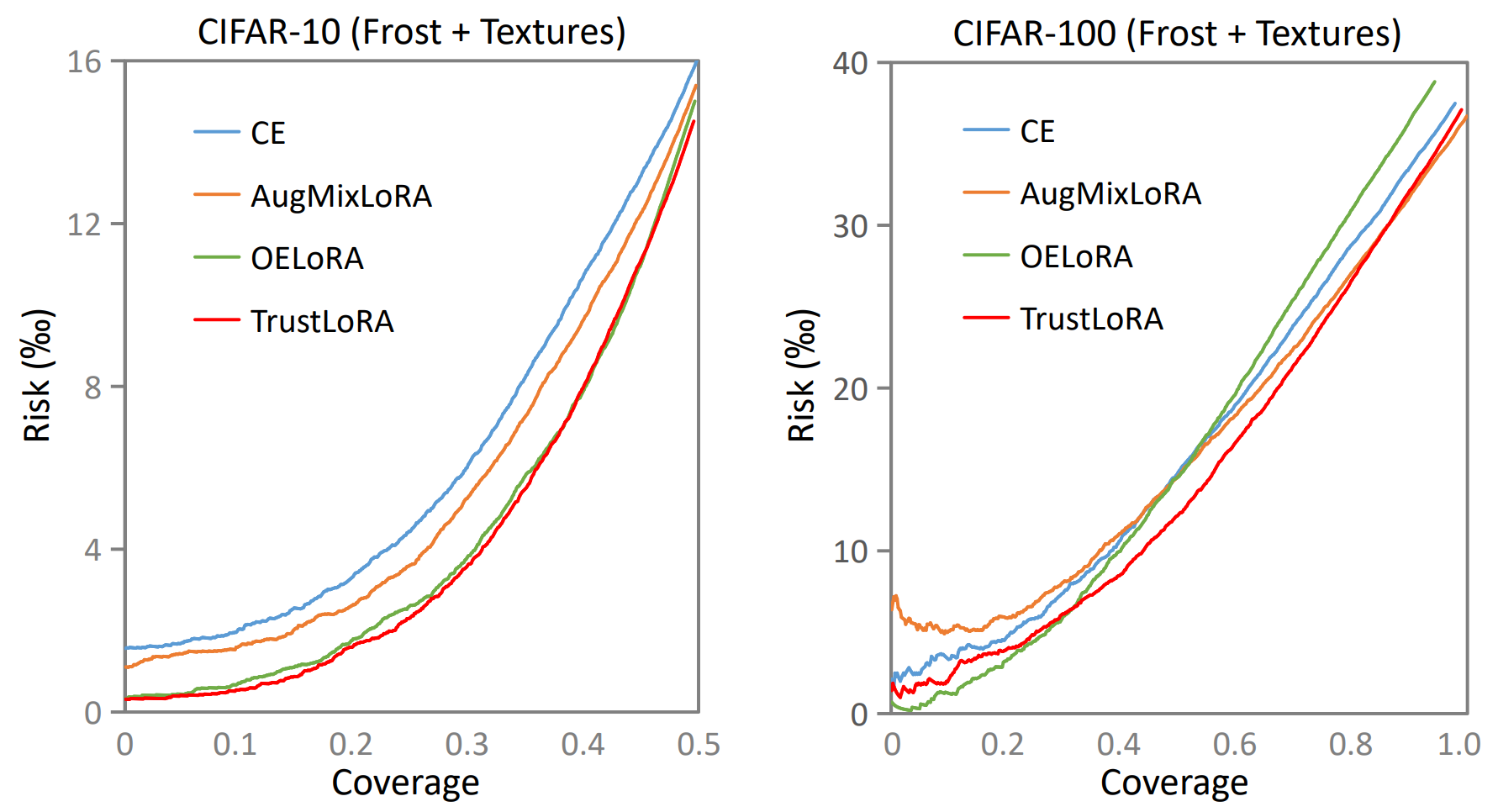}
		\vskip -0.1in
		\caption{Risk-coverage curves on the mixture of ID, covariate and semantic shifts.}
		\vskip -0.2in
		\label{figure-77}
	\end{wrapfigure}
\textbf{Risk-coverage curves.} Fig.~\ref{figure-77} provides the comparison of the risk-coverage curves when testing on the mixture of covariate shifts (Frost with Severity-1) and semantic shifts (Textures) on CIFAR10 and CIFAR-100 datasets. Ours achieves the smallest risk given a specific coverage value.

\textbf{Detailed individual failure detection performance.} Table \ref{table-10} provides the fine-grained results for misclassification detection on covariate shifts, OOD detection on semantic shifts and the unified failure detection results on C10/100 with severity-1. We observe that our method achieves the best failure detection performance. Besides, it is worth mentioning that our formulation is different from existing works that evaluate OOD generalization via accuracy and OOD detection via rejection metrics like FPR95, and AUC. We use the ARUC metric to reflect the classification with rejection ability on covariate shifts, which has integrated the classification and rejection performance. We also added the accuracy performance in Table \ref{table-10}, including methods that train the model from scratch ($^{\ast}$) and others that fine-tune the model with LoRA. All post hoc methods have the same accuracy as AugMixLoRA, since they are applied to the model trained with AugMixLoRA.

\begin{table}[t]
\caption{Individual performance on the MisD under covariate shifts, OOD detection under semantic shifts, and the unified failure detection on CIFAR-10/100.}
\vskip -0.07in
\label{table-10}
\setlength\tabcolsep{5pt}
\centering
\renewcommand{\arraystretch}{1.05}
\scalebox{0.855}{
\begin{tabular}{lcccccccccc}
\toprule[1.5pt]
\multicolumn{1}{l}{\multirow{2}{*}{{Method}}} & \multicolumn{1}{l}{\multirow{2}{*}{{ACC}}}& \multicolumn{3}{c}{{MisD-Cov}} & \multicolumn{2}{c}{{OOD Detection}} & \multicolumn{4}{c}{{Failure Detection}}\\
\cmidrule(lr){3-5} \cmidrule(lr){6-7} \cmidrule(lr){8-11}
&& {AURC} &{FPR95} &{AUC}  &{FPR95}  &{AUC}&\textbf{{AURC}} &\textbf{{FPR95}} &\textbf{{AUC}}  &\textbf{{F-AUC}} \\
\midrule[1.2pt]
& \multicolumn{10}{c}{CIFAR-10} \\
\cmidrule(lr){2-11}
CE$^{\ast}$           & 87.86 & 35.33  & 42.33 & 88.61 & 39.96  & 86.24  & 56.04  & 37.53 & 89.49 & 87.41 \\
RegMixUp$^{\ast}$           & 89.27 & 34.01  & 55.70 & 87.65 & 50.18  & 86.08  & 57.05  & 50.60 & 88.56 & 86.86 \\
CRL$^{\ast}$                & 87.68 & 29.63  & 32.91 & 90.00 & 38.08  & 86.31  & 50.25  & 32.06 & 90.35 & 88.12 \\
LogitNorm$^{\ast}$          & 87.46 & 36.45  & 40.19 & 87.57 & 20.77  & 94.46  & 48.15  & 32.56 & 91.92 & 90.88 \\
OE$^{\ast}$                 & 87.13 & 36.04  & 39.31 & 88.78 & 27.34  & 92.02  & 51.24  & 33.30 & 91.73 & 90.37 \\
RegMixUp (LoRA)     & 88.48 & 36.62  & 55.64 & 87.52 & 54.89  & 84.74  & 62.17  & 53.63 & 87.97 & 86.11 \\
CRL (LoRA)          & 88.14 & 35.04  & 43.20 & 88.40 & 41.03  & 86.36  & 56.28  & 38.40 & 89.36 & 87.37 \\
LogitNorm (LoRA)    & 86.87 & 37.83  & 38.33 & 88.76 & 40.97  & 85.74  & 59.05  & 36.71 & 89.41 & 87.22 \\
AugMix (LoRA)       & 90.53 & 18.05  & 33.51 & 90.47 & 41.92  & 87.88  & 36.62  & 36.09 & 90.72 & 89.16 \\
OE (LoRA)           & 87.38 & 32.67  & 33.82 & 89.89 & 15.47  & 96.06  & 43.63  & 26.01 & 93.51 & 92.47 \\
Energy (AugMixLoRA) & 90.53 & 26.77  & 59.77 & 84.16 & 39.36  & 89.69  & 43.50  & 49.19 & 88.03 & 86.84 \\
\rowcolor{mygray}
TrustLoRA                     & 90.77 & 16.27  & 28.98 & 91.40 & 22.41 & 93.68 & \textbf{28.68}  & \textbf{23.80} & \textbf{93.67} & \textbf{92.53} \\
\midrule
& \multicolumn{10}{c}{CIFAR-100} \\
\cmidrule(lr){2-11}
CE$^{\ast}$         & 66.08 & 134.63 & 49.00 & 84.74 & 70.45  & 69.44  & 163.66 & 54.88 & 82.95 & 76.33 \\
RegMixUp$^{\ast}$           & 68.64 & 115.84 & 48.72 & 85.10 & 69.27  & 69.84  & 154.87 & 54.34 & 83.12 & 76.72 \\
CRL$^{\ast}$                & 67.91 & 125.45 & 44.88 & 85.79 & 70.21  & 70.96  & 152.20 & 53.03 & 83.94 & 77.67 \\
LogitNorm$^{\ast}$          & 65.13 & 145.75 & 54.49 & 83.31 & 66.90  & 72.00  & 166.34 & 56.51 & 82.91 & 77.24 \\
OE$^{\ast}$                 & 62.16 & 163.93 & 51.93 & 83.08 & 47.64  & 84.68  & 149.02 & 45.48 & 87.69 & 83.87 \\
RegMixUp (LoRA)     & 67.53 & 126.07 & 50.51 & 84.54 & 66.98  & 72.39  & 155.58 & 53.51 & 83.81 & 77.99 \\
CRL (LoRA)          & 67.44 & 115.43 & 48.90 & 84.85 & 66.20  & 73.16  & 153.48 & 52.10 & 84.27 & 78.57 \\
LogitNorm (LoRA)    & 67.15 & 127.64 & 49.10 & 84.77 & 66.91  & 72.88  & 155.08 & 52.55 & 84.14 & 78.38 \\
AugMix (LoRA)       & 70.48 & 102.55 & 49.80 & 84.52 & 63.66  & 75.39  & 141.23 & 51.24 & 84.69 & 79.69 \\
OE (LoRA)           & 63.96 & 147.07 & 51.66 & 83.69 & 51.71  & 82.41  & 147.40 & 46.57 & 87.21 & 83.04 \\
Energy (AugMixLoRA) & 70.48 & 124.68 & 64.17 & 79.67 & 63.98  & 75.59  & 158.37 & 60.62 & 81.52 & 77.58 \\
\rowcolor{mygray}
TrustLoRA & 70.25 & 107.13 & 51.30 & 83.95 & 51.05 & 83.51 & \textbf{129.64} & \textbf{46.46} & \textbf{87.29} & \textbf{83.73}\\
\bottomrule[1.5pt]
\end{tabular}}
\end{table}

	\section{Conclusion}
	\vskip -0.1 in
	This work presents a novel reliability arithmetic framework to address the failure detection under both covariate and semantic shifts. For the first time, we introduce low-rank adaptation to separate and compress reliability knowledge. The proposed framework is a powerful tool to easily achieve unified, flexible and controllable reliability towards different failure sources. Extensive experiments and analysis show the superiority of our method over existing approaches for failure detection under both covariate and semantic shifts. We hope this work can inspire the community to investigate the trade-off among different failure sources, and develop flexible and controllable methods for real-world applications. 
	

	\bibliography{refer}
	
\end{document}